\documentclass{article}

\usepackage{microtype}
\usepackage{graphicx}
\usepackage{subfigure}
\usepackage{booktabs} 
\usepackage{adjustbox}
\usepackage{float}

\usepackage{hyperref}

\usepackage[accepted]{icml2023}

\usepackage{amsmath}
\usepackage{amssymb}
\usepackage{mathtools}
\usepackage{amsthm}

\usepackage[capitalize,noabbrev]{cleveref}

\theoremstyle{plain}

\theoremstyle{definition}

\theoremstyle{remark}

\usepackage[textsize=tiny]{todonotes}

\icmltitlerunning{Searching for the Essence of Adversarial Perturbations}

\begin{document}

\twocolumn[
\icmltitle{Searching for the Essence of Adversarial Perturbations}



\icmlsetsymbol{equal}{*}

\begin{icmlauthorlist}
\icmlauthor{Dennis Menn}{ntu}
\icmlauthor{Tzu-hsun Feng}{ntu}
\icmlauthor{Hung-yi Lee}{ntu}
\end{icmlauthorlist}

\icmlaffiliation{ntu}{Department of Electrical Engineering, National Taiwan University, Taipei, Taiwan}
\icmlcorrespondingauthor{Dennis Menn}{dennisymenn@gmail.com}

\icmlkeywords{Machine Learning, ICML}

\vskip 0.3in
]



\printAffiliationsAndNotice{} 

\begin{abstract}
Neural networks have demonstrated state-of-the-art performance in various machine learning fields. However, the introduction of malicious perturbations in input data, known as adversarial examples, has been shown to deceive neural network predictions. This poses potential risks for real-world applications such as autonomous driving and text identification. In order to mitigate these risks, a comprehensive understanding of the mechanisms underlying adversarial examples is essential. In this study, we demonstrate that adversarial perturbations contain human-recognizable information, which is the key conspirator responsible for a neural network's incorrect prediction, in contrast to the widely held belief that human-unidentifiable characteristics play a critical role in fooling a network. This concept of human-recognizable characteristics enables us to explain key features of adversarial perturbations, including their existence, transferability among different neural networks, and increased interpretability for adversarial training. We also uncover two unique properties of adversarial perturbations that deceive neural networks: masking and generation. Additionally, a special class, the complementary class, is identified when neural networks classify input images. The presence of human-recognizable information in adversarial perturbations allows researchers to gain insight into the working principles of neural networks and may lead to the development of techniques for detecting and defending against adversarial attacks.
\end{abstract}

\section{Introduction}
\label{submission}
Neural networks have achieved state-of-the-art performance in various machine learning tasks \cite{hinton2012deep,lecun2015deep,krizhevsky2017imagenet}. As a result, the utilization of image recognition technology has become increasingly prevalent in our daily lives. Examples of its application include text identification on license plates, facial recognition, medical image analysis, defect detection in the manufacturing industry, and autonomous vehicles. It is expected that many more applications will be developed in the near future.

Szegedy et al. have found that small perturbations incorporated with input images are sufficient to fool neural networks' judgment \cite{szegedy2013intriguing}. These small perturbations, adversarial perturbations, increase the prediction error made by neural networks, which is often done by refining adversarial perturbations iteratively through backpropagation \cite{rumelhart1986learning} to optimize a preset loss function. In practice, a number of attack algorithms have been shown to attain this goal \cite{goodfellow2014explaining, moosavi2016deepfool, papernot2016limitations, carlini2017towards, madry2017towards,  moosavi2017universal, kurakin2018adversarial}.

Similar to how a computer virus can damage or destroy a software or hardware system, an adversarial attack tricks a neural network into incorrectly categorizing an image. One potential consequence is that a self-driving car could be made to crash through an adversarial attack \cite{eykholt2018robust}. Thus, preventing adversarial attacks is crucial for the successful implementation of image recognition technology. Similar to understanding how a computer virus functions as the first step to prevent its malicious attack, understanding the working principles of adversarial attacks is crucial to countering their intended mischief. Despite the intense study of this topic since discovering adversarial perturbations in 2013 \cite{szegedy2013intriguing}, the underlying reason for their existence remains unresolved.

In contrast to previous works, we propose that adversarial perturbations contain human-recognizable information, which serves as the key conspirator responsible for neural networks' incorrect predictions. For a neural network to correctly classify an image,  it must assign high gradient values to recognizable characteristics of the image, such as the tires of a car, the wings of a plane, or the shape of an object \cite{simonyan2013deep,smilkov2017smoothgrad}. Thus, even a small alteration in pixel values of these characteristics might greatly affect the neural network's output. Consequently, we would assume that, for adversarial perturbations to confuse neural networks, changing pixel values of recognizable characteristics in an image is critical. This leads us to postulate that adversarial perturbations contain those recognizable characteristics. After all, the usefulness of a neural network is judged by us and we judge an image by recognizable characteristics.
 
Identifying human-recognizable characteristics in adversarial perturbations is challenging as these characteristics are often incomplete and corrupted by heavy noise. The reason for adversarial perturbations being noisy stems from the fact that neural networks’ gradients of the class score are noisy and perturbations are often derived from the gradients. Since the neural network does not need to utilize all the information in an image to classify the image correctly, adversarial perturbations need not modify, for example, the whole shape of a car, to fool the network. This implies that perturbations with incomplete characteristics occur naturally, contributing to the difficulty for us to identify the human-recognizable information contained in adversarial perturbations. In addition, our unfamiliarity with the working principle of neural networks, the presence of a wide variety of attack algorithms, and different attack scenarios can all contribute to the complexity of the topic. 

To support our postulation, we aim to minimize the effects of noise and reconstruct recognizable characteristics while maintaining the attack strength of adversarial perturbations. The methodology employed in this study includes obtaining perturbations from various neural networks and subsequently averaging them. This approach was chosen because noise and incomplete characteristics in perturbations generated by different neural networks are independent of one another, thus averaging would effectively minimize the impact of noise while putting together different parts of recognizable characteristics. However, obtaining a sufficient number of neural networks to average adversarial perturbations can be a limitation. To address this limitation, we adopt a technique inspired by SmoothGrad \cite{smilkov2017smoothgrad} where multiple copies of an input image incorporated with Gaussian noise are created and the adversarial perturbation for each copy is computed. Then we average the perturbations over all copies of perturbations to obtain the final perturbation.

In this study, we have evaluated our postulate in the context of different attack algorithms on the ImageNet \cite{5206848}, CIFAR10 \cite{krizhevsky2009learning}, and MNIST \cite{deng2012mnist} datasets for both targeted and untargeted attack modes. Our results indicate that human-recognizable information emerges when processed through Eqn.~\ref{gen_perturb} (to be shown below) for different attack algorithms while maintaining the ability to deceive neural networks. We have also quantitatively assessed the degree of recognizability of adversarial perturbations and made several noteworthy discoveries. Our postulate effectively explains key features of adversarial perturbations and can be used to develop countermeasures against adversarial attacks on image recognition systems.

Due to space limitations, this paper primarily presents results for ImageNet, but similar findings are seen in other datasets.

\section{Related Work}
Several previous works have attempted to unveil the nature of adversarial perturbations, yet a consensus is still lacking. Those works include:

(1) The over-linearity of neural networks \cite{goodfellow2014explaining, tramer2017space}\\
It has been shown that neural networks exhibit a linear relationship between the change in input and the corresponding change in output values. This linearity behavior, in combination with the high dimensionality of input space, makes it possible for small input changes to have a substantial effect on the network's output. This can lead to the network making inaccurate predictions when presented with adversarial examples.

(2) The presence of adversarial examples outside of the data manifold \cite{tanay2016boundary, samangouei2018defense, song2018pixeldefend}\\
Adversarial examples occur when the decision boundaries of a model extend beyond the underlying distribution of the data and are in close proximity to the input data. The goal of adversarial perturbation is to alter the input data so that it deviates from its typical distribution, causing it to cross the decision boundaries of the network and leading to incorrect predictions.

(3) Adversarial perturbations modifying non-robust features \cite{schmidt2018adversarially, tsipras2018robustness, ilyas2019adversarial}\\
The aim of adversarial perturbations is to alter non-robust features, which are derived from the data distribution, making them highly predictive for neural networks but not understandable to humans.

(4) Identifying fundamental bounds on the susceptibility of a classifier \cite{schmidt2018adversarially, shafahi2018adversarial}\\
Those works explore the possibility of preventing adversarial attacks from a theoretical perspective. Additionally, they identify fundamental bounds on the susceptibility of a classifier.

Additionally, several intriguing features linked to adversarial perturbations have also been uncovered. Four prominent features are \cite{akhtar2021advances}: 

(1) The vulnerability of neural networks to adversarial perturbations: Though the magnitude of perturbations is frequently small and not readily noticeable to the human eye when incorporated into images, adversarial perturbations have the capability to severely degrade the performance of neural networks \cite{szegedy2013intriguing}. 

(2) Transferability: It is surprising to observe that perturbations produced from one neural network can often fool networks with different architectures and even those that have been trained on different datasets \cite{szegedy2013intriguing, papernot2016transferability}. This phenomenon is known as the transferability of adversarial perturbations. Furthermore, it has been shown that these perturbations can also fool conventional machine-learning algorithms, such as support vector machines, logistic regression, and decision tree \cite{papernot2016transferability}. This highlights the importance of developing robust and secure machine learning models, particularly in sensitive applications. 

(3) Adversarial training enhancing the interpretability of neural networks: Adversarial training, a technique that trains neural networks using adversarial examples, improves the interpretability of gradients and aligns the produced adversarial perturbations with human perception, making them more understandable \cite{goodfellow2014explaining, ross2018improving, tsipras2018robustness, santurkar2019image}. This is important as the gradients of the score of the labeled class and generated adversarial perturbations in neural networks are often difficult for humans to comprehend \cite{simonyan2013deep, goodfellow2014explaining}.

(4) Non-trivial accuracy for classifiers trained on a perturbed dataset: A neural network, being trained on a dataset consisting of images that have been specifically manipulated with targeted adversarial perturbations and then relabeled to match the desired target class, has been found to exhibit an impressive accuracy of up to 63.3\% when tested on CIFAR10's original testing set \cite{ilyas2019adversarial}.

Understanding the existence of adversarial perturbations and above features not only benefits in devising effective defensive algorithms but also sheds light on the working principles of neural networks, which are often regarded as a black box.

\section{Experimental Method}
In the experiment, we utilize neural networks to generate and assess adversarial perturbations. These networks are classified into two groups: source models, which are used to create adversarial perturbations, and testing models, which are used for evaluating the attack strength and perceptibility of perturbations. 

We obtained adversarial perturbations through two different settings: (1) The single-model setting (SM) where adversarial perturbations are produced by sending an image to a source model, ResNet50 \cite{he2016deep}, and obtaining the corresponding perturbations. (2) The multiple-models with Gaussian noise (MM+G) setting where perturbations are generated according to Eqn. \ref{gen_perturb}. We also tested the single-model setting with Gaussian noise (SM+G) to observe the phenomenon of checkerboard artifacts, as described in Section ~\ref{Sec:observation}. 

In the MM+G setting, we incorporate Gaussian noise $N(0, \sigma^2 )$, with mean of $0$ and standard deviation $\sigma$, to $n$ copies of the input image $x$ and calculate the adversarial perturbations $V_i$, for the $i^{th}$ source model. Then we average the perturbations over $n$ copies and a total of $m$ sources models. Mathematically, this is

\begin{equation}
\label{gen_perturb}
	V(x) = \frac{1}{mn} \sum_{i=1}^{m} \sum_{j=1}^{n} V_{i} (x+N_{j}(0,\sigma^2))
\end{equation}

Gaussian noise is utilized with a standard deviation of 0.02 for the BIM attack and 0.05 for both the CW and DeepFool attacks. To incorporate Gaussian noise, 10 copies of each image are employed for all attack algorithms per source model, resulting in a total of 2700 produced perturbations. These perturbations are then averaged into one final perturbation.

In our ImageNet experiment, we obtained all models from PyTorchCV \cite{OlegPyTorchCV}. The testing models employed include VGG16 \cite{simonyan2014very}, ResNet50 \cite{he2016deep}, DenseNet121 \cite{huang2017densely}, and BN-Inception \cite{ioffe2015batch}. In the MM+G setting, we utilize 270 source models that are entirely distinct from testing models. Under the SM approach, the source model ResNet50 is the same as that in testing models, commonly referred as a white-box attack. We have selected 20 diverse classes, as listed in Appendix, using the first 10 images from each class in the ImageNet validation set, to give a total of 200 images and obtained respective adversarial perturbations under the SM and MM+G settings. 

We have examined BIM attack \cite{kurakin2018adversarial} and CW attack \cite{carlini2017towards} for both targeted and untargeted attack mode. Since the DeepFool attack \cite{moosavi2016deepfool} does not have a targeted mode, we only examined its untargeted attack mode. Regarding the parameters of the BIM attack, we set the value of $\epsilon$ to 0.02, $\alpha$ to 0.0008, and the number of iterations to 50. For the CW attack, we set the value of c to 1, $\kappa$ to 0, and the number of iterations to 1000. In the case of the DeepFool attack, we set the overshoot to 0.02 and the number of iterations to 50. To limit the computation time, we modified the DeepFool attack by selecting the target label from the top 10 classes with the highest output logits, rather than all classes in ImageNet. The attack implementation is based on Torchattacks \cite{kim2020torchattacks}. 

To visualize the adversarial perturbations in untargeted attack mode, we first multiply the perturbations by -1. Next, we scale the mean and variance of the perturbations to match that of the dataset. The reason for multiplying -1 is because of the effect of masking in adversarial perturbations (as described below).  Detailed information about the experimental setup is listed in Appendix.

\section{Experimental Result}
\subsection{Properties of Adversarial Perturbations}
Two distinctive properties of adversarial perturbations have been observed in this work. The first resembles the negative of recognizable characteristics in an input image. Incorporating this property with the image results in the subtraction of the pixel values of recognizable characteristics from the image, thus lowering the value of the inner product between the gradient of the labeled-class score and the input image. We name this property as masking. The second property relates to the emergence of new characteristics in adversarial perturbations. The incorporation of new characteristics into an image can make both humans and neural networks wrongly classify the image. We name this property as generation.

Masking is easier to observe compared to generation because masking refers to existing characteristics in an image whereas features produced by generation do not necessarily follow the image’s profile. As an example, a car’s shape and tires are critical for identifying its class, so adversarial perturbations from different neural networks will show similar masking features. However, there is more than one way to change a car into a plane by adding different features to the car. Thus, adversarial perturbations from different neural networks would show different features of generation. Generation is most commonly associated with targeted attacks, while masking appears in both targeted and untargeted attacks. We will illustrate the effects of masking and generation below. 

\subsection{Untargeted Attack}
\begin{figure*}[ht]
\begin{center}
\centerline{\includegraphics[width=\textwidth]{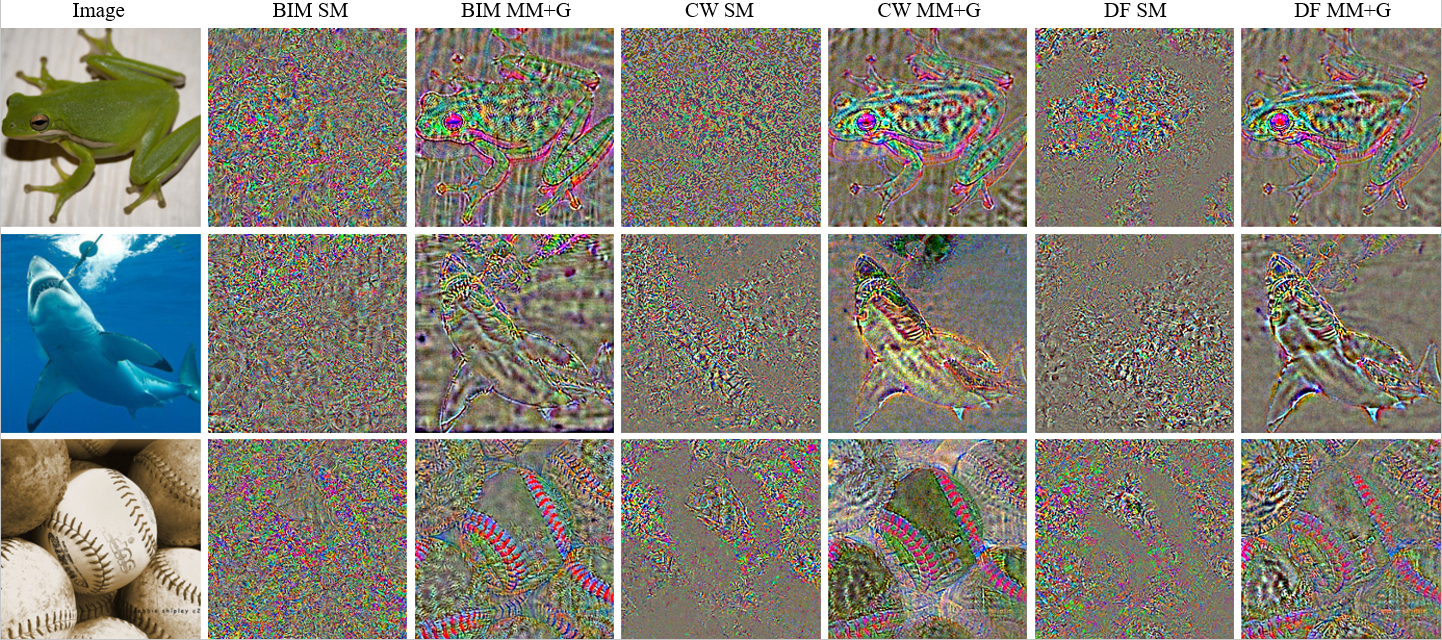}}
\caption{Adversarial perturbations generated in untargeted attack mode for images from the ImageNet dataset under the SM and MM+G settings. Under the MM+G setting, distinct and recognizable characteristics are observed in adversarial perturbations. Perturbations from CW attacks under the MM+G setting exhibit well-defined contours and minimal background noise. Despite their small magnitude, these perturbations are able to mislead neural networks' classification, highlighting the impact of masking. To enhance the visibility of the perturbations in this figure, we inverted them by multiplying by -1 and adjusted the mean and variance to match those of the dataset. Only three for a total of a thousand classes in the dataset are shown.}
\label{untargeted_all}
\end{center}
\vskip -0.3in
\end{figure*}

Figure~\ref{untargeted_all} shows input images from ImageNet and displays the negatives of the perturbations from the SM and MM+G settings for BIM, CW, and DeepFool attack algorithms. The process of generating perturbations with MM+G, compared with those with SM, minimizes the background noise and reconstructs the human-recognizable characteristics efficiently. Furthermore, perturbations with clear characteristics and low noise background produced by CW attack under the MM+G setting demonstrate the ability to attack neural networks (to be discussed in Table~\ref{attack_strength}). This suggests that masking, not noise, is key to fooling neural networks.

\subsubsection{Evaluating Recognizability}
To measure how recognizable the effect of masking in adversarial perturbations really is, we first obtain adversarial perturbations under SM and MM+G settings. We scale the values of adversarial perturbations, the same as how we scale adversarial perturbations for the purpose of visualization. Then we multiply the adversarial perturbations by 0.5 to enhance the testing model’s classification accuracy. To evaluate the classification accuracy, we set the label of each perturbation to be identical to that of the input image, as we expect the effect of masking to resemble the input images. We then send these perturbations to VGG16 for classification, noting that this model has not been included in the source models. VGG16 can achieve maximum accuracy of 56\% for predicting the labeled classes of input images when focusing on the output values of the 20 selected classes mentioned in the experimental setup. This confirms that masking resembles the recognizable characteristics in the input image, as shown in Table ~\ref{q_recogn}. Note that there is only 5\% accuracy, i.e., random guessing, under the SM setting. 

\begin{table}[ht]
\caption{Recognizability measurement for ImageNet. Under the SM setting, VGG16's classification accuracy is similar to that of random guessing, 5\%. Under the MM+G setting, the testing model’s accuracy can reach 56\%, which indicates that the recognizable characteristics in perturbations are indeed recognizable to neural networks.}
\label{q_recogn}
\vskip 0.05in
\begin{center}
\begin{small}
\begin{sc}
\begin{tabular}{lcccr}
\toprule
Attack Algorithms & SM & MM+G \\
\midrule
BIM       & 5.5\%& 56.0\% \\
CW        & 4.5\%& 38.0\% \\
DeepFool  & 5.0\%& 38.0\% \\
\bottomrule
\end{tabular}
\end{sc}
\end{small}
\end{center}
\vskip 0.05in
\end{table}

To assess the human perceptibility of masking in adversarial perturbations, we conducted a human evaluation test. We generated 200 perturbations using BIM attacks under the MM+G setting, a procedure identical to those of the previous experiment. The perturbations were scaled for visualization, as described earlier. We then randomly divided the 200 perturbations into four equal groups. Twelve people were tasked with determining the most appropriate label for each group, using the 20 selected classes specified in the experimental setup. After removing the minimum and maximum classification accuracy for each group, we achieved an average accuracy of 80.7\%, indicating that masking in adversarial perturbations is highly visible for humans. Random guessing would yield 5\% accuracy.

\subsubsection{Evaluating Attack Strength}
To verify if adversarial perturbations obtained from the MM+G setting are still capable of attacking neural networks, we evaluate their attack strength. We have produced adversarial perturbations from the 200 selected images under SM and MM+G settings. Then, we processed the perturbations by multiplying a quantity $\varepsilon$ with their signs, as shown below:

\begin{equation}
\label{attack_eval}
	V' = \varepsilon \cdot sign( V )
\end{equation}

\begin{figure}[hbt]
\vskip 0.2in
\begin{flushleft}
\centerline{\includegraphics[width=\columnwidth]{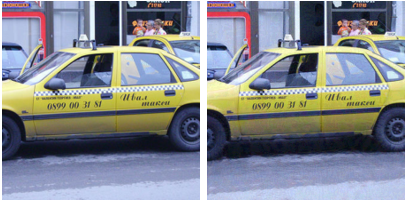}}
\hskip 4em Image \hskip 7em  Adversarial example
\caption{Incorporation of adversarial perturbations. The input image from ImageNet has been modified by incorporating adversarial perturbations generated by the BIM attack. The perturbation was processed by Eqn.~\ref{attack_strength} with $\epsilon$ = 0.02. While the adversarial perturbation may not be easily discernible to the human eye, it is effective in misleading the neural network.}
\label{pic:adv_img}
\end{flushleft}
\vskip -0.1in
\end{figure}

Here $V$ and $V’$ are the original and processed adversarial perturbations, respectively. $\varepsilon$ = 0.02 for all attack algorithms. This process is similar to the fast gradient sign method \cite{goodfellow2014explaining}. As a result, every perturbation has the same $L_{inf}$ and $L_{2}$ norm. Next, we incorporated the processed perturbations with the input images and sent them to testing models to evaluate their classification accuracy. An adversarial example, as generated by Eqn.~\ref{attack_eval} with $\epsilon$ equals 0.02, is depicted in Figure~\ref{pic:adv_img}. The figure demonstrates an input image and adversarial example produced by the BIM attack. Although the differences between the input image and the adversarial example may not be readily noticeable to the human eye, they are sufficient to deceive testing models into misclassifying a cab as an ambulance.

The attack strengths of the adversarial perturbations obtained from different attack algorithms for ImageNet are listed in Table \ref{attack_strength}. The Image column of Table~\ref{attack_strength} shows that, on average, the testing models correctly classify 81.8\% of the input images. The effect of adding noise (Noise column), defined as a random sampling where each pixel value has an equal probability of being +0.02 or -0.02, to images does not affect testing models' decision. The incorporation of perturbations under the SM setting makes the classification accuracy drop, on average, to 63.3\% for the BIM attack. Perturbations from MM+G further reduce the classification accuracy to 13.2\% on average for the BIM attack. This confirms the strong ability of masking in attacking neural networks. This result also applies to CIFAR10 and MNIST datasets.

\begin{table*}[hbt]
\caption{ Attack strength of adversarial perturbations processed by MM+G. Values under the “Image” column represent classification accuracies on clean images and those of “Noise” are accuracies after incorporating random noise into images. In the SM setting, both the source and testing models use the same ResNet50, this is also known as white-box attack. Values obtained from ResNet50 are excluded from calculating values in the row of “Avg.” under the SM setting.}
\label{attack_strength}
\vskip 0.15in

    \begin{center}
    \begin{small}
    \begin{sc}
    \begin{adjustbox}{max width=\textwidth}
    \begin{tabular}{lcccccccc}
    \toprule
    \multicolumn{1}{c}{Testing Models} & \multicolumn{2}{c}{Reference} & \multicolumn{3}{c}{SM}   & \multicolumn{3}{c}{MM+G} \\ 
    \midrule
                                       & Image         & Noise         & BIM    & CW     & DF     & BIM    & CW     & DF     \\ 
    BN-Inception                       & 81.5\%        & 81.5\%        & 64.0\% & 77.0\% & 68.0\% & 16.5\% & 22.0\% & 15.0\% \\
    DenseNet121                        & 83.5\%        & 83.5\%        & 58.5\% & 77.5\% & 66.5\% & 10.5\% & 16.5\% & 13.0\% \\
    VGG16                              & 79.0\%        & 79.0\%        & 67.5\% & 76.5\% & 70.5\% & 12.5\% & 20.5\% & 17.5\% \\
    ResNet50                           & 83.0\%        & 83.0\%        & 0.0\%  & 7.0\%  & 4.5\%  & 13.0\% & 18.5\% & 14.0\% \\
    \midrule
    Avg.                               & 81.8\%        & 81.8\%        & 63.3\% & 77.0\% & 68.3\% & 13.2\% & 19.7\% & 15.2\% \\
    \bottomrule
    \end{tabular}
    \end{adjustbox}
    \end{sc}
    \end{small}
    \end{center}
    \vskip 0.1in
\end{table*}

\begin{figure*}[ht]
\vskip 0.2in
\begin{flushleft}
\centerline{\includegraphics[width=\textwidth]{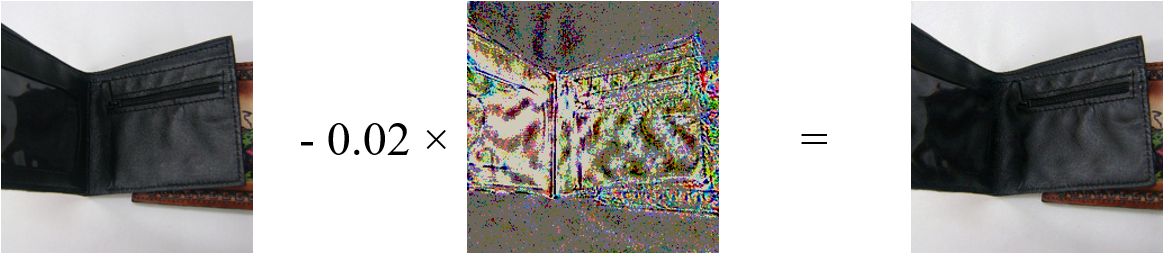}}
\ \ \ \ \ \ \ \ \ \ \ \ \ \ \ Wallet  \hskip 16em  \ Matchstick  \hskip 14em \ \  Mailbag \\
\hskip 1em \ \ \  confidence: 64.2\% \hskip 12em \  confidence: 2.9\% \hskip 11em \ confidence: 21.3\%
\caption{Effect of masking. To show that the masking property is a key factor in deceiving neural networks, we remove the background noise in adversarial perturbations produced under the MM+G setting. Despite this modification, the attack remained successful. The predicted classes and confidence scores for ResNet50's classification of the above images are shown.}
\label{demo}
\end{flushleft}
\vskip -0.2in
\end{figure*}

To further demonstrate that the masking property of adversarial perturbations is a crucial factor in deceiving neural networks, we remove background noise in adversarial perturbations under the MM+G setting by equating values with an $L_{1}$ norm smaller than the average $L_{1}$ norm in a perturbation to zero. Despite this modification, the attack remained successful to fool testing models, as illustrated in Figure ~\ref{demo}. We successfully make ResNet50 misclassify a wallet as a mailbag by incorporating the adversarial perturbation to the input image. We have also conducted experiments on CIFAR10 and obtained results similar to those obtained from the ImageNet dataset.

\subsection{Targeted Attack}
\begin{figure}[ht]
\vskip 0.2in
\begin{flushleft}
\centerline{\includegraphics[width=\columnwidth]{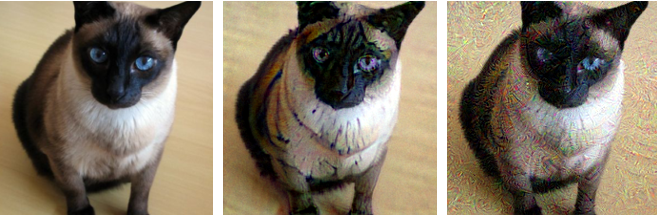}}
\hskip 2em \ \  Image \hskip 3em \ \ \ \  CW MM+G \hskip 3em \ \ CW SM
\caption{Adversarial perturbations generated in targeted attack mode under the MM+G and SM settings. In this example, the original image is labeled as Siamese cat and the targeted label for the attack is tiger. Under the MM+G setting, the tiger-like traits in the adversarial example becomes more pronounced, demonstrating the effect of generation. This observation reinforces our hypothesis that adversarial perturbations carry human-recognizable characteristics.}
\label{targeted}
\end{flushleft}
\vskip -0.2in
\end{figure}

The effect of generation in adversarial perturbations gets more pronounced in the targeted attack mode. In the targeted attack mode, the experimental setup is identical to that of the untargeted attack, except the CW attack employs a higher attack strength by setting the values of $c$ and $\kappa$ to 5. Figure ~\ref{targeted} illustrates one example of adversarial example produced by the targeted CW attack algorithm. To improve visualization, we scale the adversarial perturbation by multiplying it with a constant, so that the maximum value in the perturbation is 0.5. In this attack, the target for the Siamese cat is chosen to be a tiger. By modifying a portion of the input image, we can successfully transfer the original image to the target image. Under the MM+G setting, we observe that the color of the cat's fur is transformed to orange, the stripes become more prominent, and the eye color changes from blue to orange. These modifications correspond to the characteristics of a tiger, reinforcing our hypothesis that adversarial perturbations carry human-identifiable information. We have also observed similar results for the targeted BIM attack in the MM+G setting. 

It becomes challenging to achieve comparable results when the targeted class and labeled class are vastly dissimilar, such as when we attempt to change a cat image to a car image. There are multiple ways to accomplish a targeted attack, so different models may generate different features. Averaging all results may also cancel out any distinctive features. This would explain why targeted label manipulation is not easily transferable to different models without additional operations \cite{liu2017delving}.

\section{Discussion}
Our work has revealed that adversarial perturbations possess human-recognizable characteristics, which are responsible for deceiving neural networks. Four key features of adversarial perturbations mentioned earlier can now be understood and explained based on this finding:

(1) The vulnerability of neural networks to adversarial perturbations: For neural networks to correctly classify an input image, they put large gradient values on the recognizable characteristics when calculating the score of a labeled class. As mentioned before, a slight change in the pixel value by adversarial perturbations can greatly influence neural networks’ prediction. This is consistent with our observation that vulnerability stems from small changes in recognizable characteristics in input images. 

(2) Transferability: To get an accurate identification, the neural network must utilize some recognizable features of an input image, such as the shape of a shark, as shown in Figure~\ref{untargeted_all}. Therefore, for the purpose of classification, a significant portion of the image's recognizable characteristics likely overlap for different neural networks. If these overlapped features are altered by the incorporation of adversarial perturbations, it leads to incorrect classifications even for unrelated neural networks.

(3) Adversarial training enhancing the interpretability of neural networks: As long as the gradient of the class score remains roughly the same when incorporating perturbations to an input image \cite{goodfellow2014explaining}, adversarial training in effect minimizes $\ L_{2}$ norm of the gradient for the score of each class \cite{simon2019first}. Since noise is included in the gradient, upon the minimization of $\ L_{2}$ norm, noise will be reduced. Minimizing $\ L_{2}$ norm can also help to reproduce recognizable characteristics. We give an example to demonstrate this concept. Assuming two pixels are equally important in a classification task, the sum of the two inner product pairs (each pixel and corresponding gradient) remains the same as long as the sum of the two gradients is fixed. However, $\ L_{2}$ norm reaches a minimum if the weights distribute equally on the two pixels. So neural networks will tend to distribute the values of the gradient equally. As a result, missing or less emphasized portions in the gradient can now be reconstructed and this leads to more complete information. Thus, adversarial training is able to de-noise and reproduce gradient information similar to what we have shown earlier with MM+G, resulting in more recognizable perturbations.

(4) Non-trivial accuracy for classifiers trained on a perturbed dataset: Classifiers trained on a perturbed dataset and relabeled as the target class have demonstrated to achieve a non-trivial accuracy on a clean testing set \cite{ilyas2019adversarial}. Our postulate explains this feature: Since adversarial perturbations contain human-recognizable characteristics, neural networks trained with a perturbed dataset retain the knowledge of those recognizable characteristics and therefore are able to attain non-trivial classification accuracy in clean data.

\section{Related Observation} \label{Sec:observation}
The utilization of MM+G allows us to not only see the recognizable aspect of adversarial perturbations literally but also preserve the attack ability on neural networks. Three noteworthy observations are summarized below:

(1) Halos in adversarial perturations\\
We have placed the input image and the corresponding adversarial perturbation side by side to find differences between the two. Interestingly, we find halos enclosing the boundaries of the objects of interest have been present in many perturbations, as shown in the shark and the frog in Figure ~\ref{untargeted_all}. This phenomenon has also been observed for perturbations generated from the CIFAR10 dataset. The exact cause for the halos remains to be identified.

\begin{figure}[ht]
\vskip 0.2in
\begin{flushleft}
\centerline{\includegraphics[width=\columnwidth]{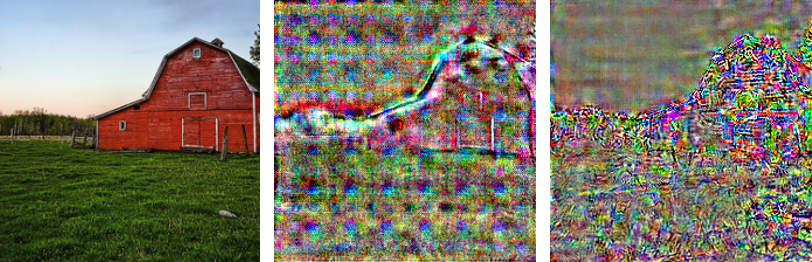}}
\hskip 2em \ \  Image \hskip 3em \ \ \ \ \ \ STD = 0.6 \hskip 3em \ STD = 0.05
\caption{Checkerboard artifacts in adversarial perturbations. 
When applying the SM+G setting to produce adversarial perturbations, checkerboard-like patterns tend to emerge. Increasing the standard deviation of the added noise results in more prominent patterns and different periods of the patterns.}
\label{std}
\end{flushleft}
\vskip -0.1in
\end{figure}

(2) Checkerboard artifacts in adversarial perturbations\\
We apply 1000 different Gaussian noises, each having a mean of 0 and standard deviation of 0.05, to each input image and sum the corresponding perturbations. We find that checkerboard-like patterns emerge, albeit not prominently. To further investigate this phenomenon, we increase the standard deviation of Gaussian noise to 0.6, resulting in more obvious patterns and altered periods, as shown in Figure ~\ref{std}. The cause of the patterns may be closely related to the study of checkerboard artifacts \cite{odena2016deconvolution}.

(3) The missing cat \\
We find that neural networks will classify random noises into specific classes, regardless of the dataset or model used. Below we present these findings using both CIFAR10 and ImageNet datasets.
 
CIFAR10:\\
We observe that the contours of adversarial perturbations for the class of cats in CIFAR10 are difficult to identify, yet the corresponding adversarial perturbations are still classified as a cat by testing models. To investigate further, we have sent 100 Gaussian noises, with mean and variance being the same as that of the CIFAR10 dataset, to four different testing models for classification. Surprisingly, the classification results highly concentrate on the classes of cat and bird, as listed below.

1. DenseNet40\\
\hspace*{2.3ex}cat: 100\%.

2. DiaResNet164\\ 
\hspace*{2.3ex}cat: 77\%, bird: 23\%.

3. PyramidNet110\\ 
\hspace*{2.3ex}cat: 95\%, bird: 5\%.

4. ResNet56\\ 
\hspace*{2.3ex}bird: 100\%.

One possible explanation is that the classes of cat and bird have a wide range of shapes and postures in the CIFAR10 training set, making it difficult for the network to identify specific characteristics. As a result, the neural network may develop a strategy of classifying images that are not easily distinguishable as either cat or bird, which we refer as the "complementary class".

ImageNet:\\
We send 1000 Gaussian noises, matched for mean and variance to that of the ImageNet dataset, to the testing models for classification. The results show a high concentration of classification on chain mail, prayer rug, and doormat for BN-Inception, DenseNet121, and VGG16 models, respectively, as shown below:

1. BN-Inception\\
\hspace*{2.3ex}chain mail: 99.3\%, prayer rug: 0.3\%, doormat: 0.2\%,\\
\hspace*{2.3ex}chain: 0.2\%.

2. DenseNet121\\
\hspace*{2.3ex}prayer rug: 93.6\%, chain mail: 6.0\%, nematode: 0.4\%.

3. VGG16\\
\hspace*{2.3ex}doormat: 98.9\%, sandbar: 1.0\%, dishrag: 0.1\%.

4. ResNet50\\
\hspace*{2.3ex}window screen: 78.8\%, jean: 13.7\%, tennis ball: 3.8\%,\\
\hspace*{2.3ex}armadillo: 2.8\%, golf ball: 0.8\%, goose: 0.1\%.

The classes of prayer rug, chain mail, doormat, and window screen all contain fundamental geometric shapes such as grids and squares. Our hypothesis is that the neural network uses these base classes, containing only fundamental shapes, as a starting point for image classification. Without additional information, the neural network's prediction will stay with the base classes, which become the complementary classes in ImageNet.

\section{Conclusion}
Previous research has uncovered a discrepancy between human perception and neural network classification when it comes to adversarial perturbations. Our study shows that these perturbations possess human-recognizable characteristics, which lead to neural network's misclassification. Our investigation suggests that adversarial perturbations, for both targeted and untargeted attacks, are human recognizable and can be utilized as a means to understand neural network's judgment.

Furthermore, in this work we have identified two properties in adversarial perturbations: masking and generation. We have also presented several intriguing findings: the presence of a halo surrounding adversarial perturbation, the existence of checkerboard artifacts in adversarial perturbations, and the discovery of the missing cat which leads to the identification of the complementary class. By using MM+G, we can dig deeper into the functioning of neural networks and gain insights that can aid in the development of defense against adversarial attacks.


\bibliography{example_paper}
\bibliographystyle{icml2023}

\newpage
\appendix
\onecolumn

\section {Detailed Information for Attack Setting}
\subsection{Selected 20 Classes in the Experimental Method}
The 20 classes utilized from the ImageNet dataset are listed below:

great white shark, cock, tree frog, green mamba, giant panda, ambulance, barn, baseball, broom, bullet train, cab, cannon, teapot, teddy, trolleybus, wallet, lemon, pizza, cup, daisy.

\subsection{Calibrating Attack Algorithms}
Incorporating Gaussian noise with input images can affect the class scores produced by a neural network, diminishing the effectiveness of adversarial perturbations and increasing the likelihood of misclassification of the input image. In some algorithms, this misclassification can also lead to the failure of algorithm initiation. To mitigate the impact of incorporated noise on class scores and prevent misclassification, we calibrate the class scores to match those obtained in the absence of noise before launching any attacks on the images. Mathematically, this means

  \begin{equation}
  \label{attack_mod}
  \begin{aligned}
        &M'(x+N(0,\sigma^{2})+V(x)) = M(x+N(0,\sigma^{2})+V(x)))+calibration\\
        &calibration = M(x) - M(x+N(0,\sigma^{2}))
  \end{aligned}
\end{equation}

where $M(x)$ and $M’(x)$ are the original and the calibrated class score for a given image $x$, respectively; $N(0,\sigma^{2})$ is Gaussian noise with mean $0$ and variance $\sigma^{2}$, and $V(x)$ is the adversarial perturbations calculated from the input image $x$.

\subsection{Effect of Clipping}
A standard procedure for processing adversarial examples is to clip the pixel values to the range between 0 and 1. This
procedure is not followed for adversarial perturbations $V_{i}$ in Eqn.~\ref{gen_perturb} because clipping will affect the perceptibility of perturbations.


\end{document}